\documentclass{article} % For LaTeX2e
\usepackage{iclr2026_conference,times}

\usepackage{amsmath,amsfonts,bm}

\def\eqref#1{equation~\ref{#1}}
\def\1{\bm{1}}

\def\vzero{{\bm{0}}}
\def\vone{{\bm{1}}}
\def\vmu{{\bm{\mu}}}
\def\vsigma{{\bm{\sigma}}}

\DeclareMathAlphabet{\mathsfit}{\encodingdefault}{\sfdefault}{m}{sl}
\SetMathAlphabet{\mathsfit}{bold}{\encodingdefault}{\sfdefault}{bx}{n}

\usepackage{hyperref}
\usepackage{url}
\usepackage{graphicx} % For including images
\usepackage{booktabs} % For professional looking tables
\usepackage{amsmath}  % For \odot
\usepackage{subcaption}
\usepackage{algorithm}
\usepackage{algpseudocode}
\usepackage{setspace}
\usepackage{footnote}
\usepackage{soul}
\usepackage{xcolor}

\title{Warm Starts Accelerate Conditional Diffusion}

\author{Jonas Scholz \\
University of Cambridge\\
Cambridge, UK\\
\texttt{js2731@cam.ac.uk}
\And Richard E. Turner \\
University of Cambridge\\
Cambridge, UK\\
\texttt{ret26@cam.ac.uk}
}

\iclrfinalcopy % Uncomment for camera-ready version, but NOT for submission.
\begin{document}

\maketitle

\begin{abstract}
Generative models like diffusion and flow-matching create high-fidelity samples by progressively refining noise. The refinement process is notoriously slow, often requiring hundreds of function evaluations.
We introduce \emph{Warm-Start Diffusion} (WSD), a method that uses a simple, deterministic model to dramatically accelerate \emph{conditional} generation by providing a better starting point.
Instead of starting generation from an uninformed $\mathcal{N}(\vzero, I)$ prior, our deterministic warm-start model predicts an informed prior $\mathcal{N}(\hat{\vmu}_C, \text{diag}(\hat{\vsigma}^2_C))$, whose moments are conditioned on the input context $C$. This \emph{warm start} substantially reduces the distance the generative process must traverse, and therefore the number of diffusion steps required, particularly when the context $C$ is strongly informative. WSD is applicable to any standard diffusion or flow matching algorithm, is orthogonal to and synergistic with other fast sampling techniques like efficient solvers, and is simple to implement. We test WSD in a variety of settings, and find that it substantially outperforms standard diffusion in the efficient sampling regime, generating realistic samples using only 4-6 function evaluations, and saturating performance with 10-12.
\end{abstract}

\section{Introduction}

Generative models based on stochastic processes, like diffusion and flow-matching, have become the state-of-the-art for high-fidelity data synthesis \citep{ho_denoising_2020, song_score-based_2020, karras_elucidating_2022}. 
Although diffusion can be used to generate samples using very little conditioning information (e.g.~text-to-image generation) or no conditioning information at all (unconditional diffusion), many domains rely on highly informative context information $C$ to guide generation. For instance:
\begin{itemize}
    \item Image inpainting, super-resolution, noise-removal, and colouration ($C = $ available pixels)
    %\item Image super-resolution ($C = $ low-resolution image)
    \item Video and audio generation ($C =$ previous frames or spectral coefficients)
    \item Molecule generation ($C = $ molecule properties \citep{3d_molecule_diffusion} or graph of atoms \citep{xugeodiff})
    \item Weather forecasting ($C =$ current weather) \citep{kong_diffwave_2021, ho_imagen_2022, price_gencast_2024}
    \item Fluid dynamics simulators ($C=$ previous state) \citep{shu2023physics}
\end{itemize}
Despite the success of diffusion in these domains, its practical application is often limited by a significant bottleneck: slow, iterative sampling that can require a Number of Function Evaluations (NFE) in the hundreds to generate a single sample. This cost becomes particularly problematic in domains where each sample is itself only part of an autoregressive rollout that can contain hundreds or thousands of samples, highlighting the importance of computationally efficient methods for conditional diffusion. Our work focuses on accelerating sampling for this class of problems.

Significant progress has been made from the inefficient foundational DDPM method \citep{ho_denoising_2020} that required $\sim 1000$ steps per sample: Re-framing the diffusion process in a continuous-time setting opened the door for much faster sampling \citep{song_score-based_2020}. Subsequent methods have further reduced the step count by developing more efficient ways to solve the underlying ordinary differential equation (ODE). These advancements include deterministic samplers like DDIM \citep{song_denoising_2022}, which enabled larger step sizes; higher-order numerical solvers like DPM-Solver(++) \citep{lu_dpm-solver_2022, lu2025dpmpp}, which approximate the ODE solution more accurately per step; and novel training paradigms like flow matching \citep{lipman_flow_2022}, which aim to learn simpler, straighter generative paths that are inherently easier to integrate. Combining these advanced techniques, high-quality samples can now be generated in tens of sampling steps.

Conceptually, all of these methods reduce the number of sampling steps by increasing the \emph{distance covered by each sampling step}, allowing for fewer, larger steps to reach the data distribution.
In this work, we instead propose a method that reduces the \emph{total distance} to be traversed in the first place by moving the initial distribution closer to the data distribution, based on the context information $C$.

\begin{figure}[t]
\centering
\includegraphics[width=\linewidth]{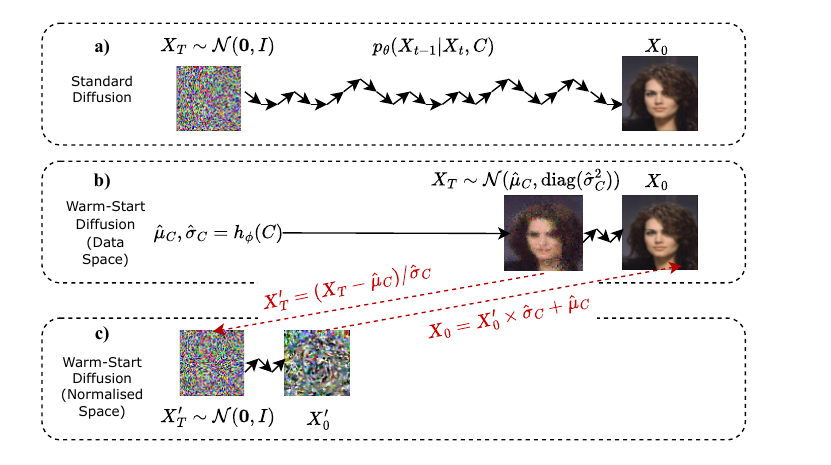}
\caption{\textbf{a)} In standard diffusion, many steps are needed to transform a sample $X_T \sim \mathcal{N}(\vzero, I)$ to $X_0 \mid C \sim p(X_0 \mid C)$. \textbf{b)} Using a warm-start model $h_{\phi}$, we can draw an initial sample $X_T \mid C \sim \mathcal{N}(\hat{\vmu}_C, \text{diag}(\hat{\vsigma}^2_C))$ that is already close to the data distribution, allowing us to traverse the gap in fewer steps. %However, the generative model's algorithm would need to be adjusted to account for the non-standard noise distribution.
\textbf{c)} By working in an equivalent sample-normalised space, where $X_T' \sim \mathcal{N}(\vzero, I)$, a normalised-space sample $X'_0 \mid C$ can be drawn using standard diffusion, and is then unnormalised to obtain a sample $X_0 \mid C$ from the data distribution.}
\label{fig:warm-start-models}
\end{figure}

Other generative methods that are fast at inference time exist, but each has its own shortcomings: GANs \citep{goodfellow2020generative} can generate images in a single forward pass but are difficult to train and can suffer from mode collapse. Consistency models \citep{song2023consistency} are modern alternatives, but require the complex and brittle distillation of a pre-trained diffusion model. In the domain of weather forecasting, single-step generative models relying on the Continuous Ranked Probability Score (CRPS) have shown recent success \citep{lang2024aifscrps, alet2025skillful}, but this method is domain-specific and potential shortcomings are not yet fully understood\footnote{For instance, as the CRPS only considers marginal distributions, the loss does not inherently guarantee realistic joint distributions.}.

In summary, our contributions include:
\begin{itemize}
    \item The warm-start diffusion approach, which substantially reduces the computational cost of sampling in conditional diffusion settings.
    \item A conditional normalisation trick, that makes our method compatible with any standard diffusion framework, and easy to implement.
    \item A detailed evaluation on image inpainting and weather forecasting tasks demonstrating the method's effectiveness.
    \item A discussion of the limitations of this method, particularly with regard to unconditional or weakly conditional diffusion domains.
\end{itemize}

\section{Warm-Start Diffusion}
\label{sec:warm-start-models}
Our main contribution is \emph{Warm-Start Diffusion} (WSD) --- a method that speeds up sampling in conditional diffusion by moving the noise distribution closer to the data distribution. Instead of drawing the initial noise sample $X_T$ from a standard normal distribution $ X_T \sim \mathcal{N}(\vzero, I)$, WSD uses a small, deterministic \emph{warm-start model} to predict a conditional mean $\hat{\vmu}_C$ and marginal standard deviation $\hat{\vsigma}_C$ from a given context $C$.
Using these moments, a noisy sample can be drawn from the \emph{informed} prior $p(X_T \mid C) = \mathcal{N}(\hat{\vmu}_C, \text{diag}(\hat{\vsigma}^2_C)$, which we write as $\mathcal{N}(\hat{\vmu}_C, \hat{\vsigma}_C)$ for brevity. By using this informed prior as the starting point for an entirely separate generative model, we can skip a large number of initial sampling steps. This is illustrated in Fig. \ref{fig:warm-start-models}.

We adopt the DDPM notation, where $t \in [0, T]$ defines a timestep in the sampling process, with $t=0$ being the final sample from the data distribution and $t=T$ being the initial noise sample.

\subsection{Generation}
\label{sec:inference}
The full generative process requires three components:
\begin{itemize}
    \item Context data $C$ (e.g.~fixed pixels in an inpainting task, or the current weather in a weather forecasting task).
    \item A warm-start model $h_{\phi}$ that takes the context data $C$ and outputs the first two moments of the conditional data distribution $p(X_0 \mid C)$, i.e.~the mean and marginal standard deviation $\hat{\vmu}_C$ and $\hat{\vsigma}_C$.
    \item A generative model\footnote{Here, $p_{\theta}$ is implemented by an iterative solver. When using a deterministic ODE solver, this conditional distribution is a Dirac delta.} $p_{\theta} (X_0 \mid X_T, C, \hat{\vmu}_C, \hat{\vsigma}_C)$, that generates samples from the conditional data distribution $p(X_0 \mid C)$, given the context data $C$ and a noise sample $X_T \sim \mathcal{N}(\hat{\vmu}_C, \hat{\vsigma}_C)$.
\end{itemize}
An explanation of how $h_{\phi}$ and $p_{\theta}$ can be obtained is given in Section~\ref{sec:training}.

\begin{figure}[t]
\centering
\includegraphics[width=\textwidth]{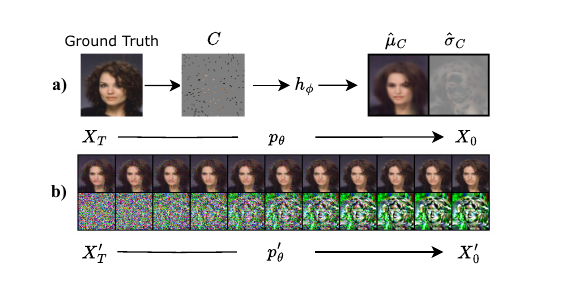}
\caption{The entire 10-step sampling process for image inpainting. \textbf{a)} The context data $C$ is a masked ground truth image with 5\% of the pixels visible. The warm-start model $h_{\phi}$ predicts a conditional mean and marginal standard deviation. \textbf{b)} By starting with a sample from $\mathcal{N}(\hat{\vmu}_C, \hat{\vsigma}_C)$ and applying standard diffusion, a realistic sample $X_0$ is generated. The bottom row shows the same process but in normalised space, where $X'_T \sim \mathcal{N}(\vzero, I)$.}
\label{fig:inpainting-samples}
\end{figure}

The process to generate a sample $X_0$ from context $C$ is:
\begin{equation}
\hat{\vmu}_C, \hat{\vsigma}_C = h_{\phi}(C), \quad X_T \sim \mathcal{N}(\hat{\vmu}_C, \hat{\vsigma}_C), \quad X_0 \sim p_{\theta}(X_0 \mid X_T, C, \hat{\vmu}_C, \hat{\vsigma}_C),
\end{equation}
which is shown in Figs. \ref{fig:warm-start-models} and \ref{fig:inpainting-samples}.

\subsection{The Conditional Normalisation Trick}
\label{sec: conditional normalisation}
Many common diffusion algorithms are derived with the assumption that noise is sampled from a \textit{standard} Gaussian $X_T \sim \mathcal{N}(\vzero, I)$. To make these diffusion algorithms compatible with WSD, where $X_T \sim \mathcal{N}(\hat{\vmu}_C, \hat{\vsigma}_C)$, they would potentially need to be re-derived and re-implemented. We sidestep this inconvenience using the conditional normalisation trick.

It is well known that the base distribution $\mathcal{N}(\hat{\vmu}_C, \hat{\vsigma}_C)$ can be shifted by $\hat{\vmu}_C$ and scaled by $\hat{\vsigma}_C$ to produce a standard normal $\mathcal{N}(\vzero, I)$. If we apply the same transformation on a per-instance basis to all steps of the diffusion process $X_t$, the generative model can perform diffusion in an instance-normalised space, $X'_t$:
\begin{equation}
    \label{eq: normalisation}
    X_t \to X'_t = (X_t - \hat{\vmu}_C) / \hat{\vsigma}_C.
\end{equation}
Intuitively, in data space, WSD moves the noise distribution closer to the data distribution. In normalised space, WSD moves \emph{the data distribution closer to the noise distribution}, by \emph{removing} the first two moments from the data distribution. Both approaches are mathematically equivalent, but the latter allows for significantly easier implementation because $X'_T \sim \mathcal{N}(\vzero, I)$, recovering the standard diffusion assumption. Both are shown in Figs.~\ref{fig:warm-start-models} and \ref{fig:inpainting-samples}.
Generation in normalised space thus becomes: \vspace{1pt}
\begin{equation}
X'_T \sim \mathcal{N}(\vzero, I), \quad 
X'_0 \sim p'_{\theta} (X'_0 \mid X'_T, C, \hat{\vmu}_C, \hat{\vsigma}_C), \quad
X_0 = X'_0 \cdot \hat{\vsigma}_C + \hat{\vmu}_C. \label{eq: unnormalisation}
\end{equation}
\vspace{1pt}In Sec. \ref{sec:training} and Alg. \ref{alg:warm-start-training}, we explain how $p'_\theta$ is trained. 

\subsection{Warmth Blending and Multi-Task Training} \label{sec: warmth blending}
We find that WSD significantly improves image quality for low NFE. However, in the large NFE regime, standard flow matching performs better. This is shown as an ablation in Fig.~ \ref{fig:inpainting_nfe_vs_fid} (right, red). We hypothesise that this underperformance is related to $\hat{\vsigma}_C$: In regions where the warm-start model is very confident (and $\hat{\vsigma}_C$ small), it acts as an overly strong constraint that might inhibit the generative model's performance.

We overcome this limitation by introducing multi-task training that includes a range of diffusion tasks, ranging from WSD to standard diffusion. Specifically, we introduce the \textit{warmth}, $w$, and modify $\hat{\vsigma}_C$ so that
\begin{equation}
    \hat{\vsigma}_C^{(\text{norm})} = w \cdot \max (\hat{\vsigma}_C, 1 - w) + (1 - w) \vone
\end{equation}
is used for (un)normalisation. We also pass $w$ to $p'_\theta$ as an additional scalar input. This means that for $w = 0$, $\hat{\vsigma}_C^{(\text{norm})} = \vone$, and for $w = 1, \hat{\vsigma}_C^{(\text{norm})} = \hat{\vsigma}_C$. Effectively, $w$ blends the standard and ``warm" diffusion tasks, which we find improves performance. During training (Alg. \ref{alg:warm-start-training}), $w$ is randomly sampled $w \sim \mathrm{U}[0, 1]$. During inference (Alg. \ref{alg:warm-start-sampling}), $w$ is a hyperparameter, which we simply set to 1 for all experiments\footnote{We find that using slightly smaller values of $w = 0.8$ in the high NFE regime yields very slightly better FID scores, but find these gains to be visually imperceptible and not worth the additional complexity of adapting $w$.}.

\subsection{Training}
\label{sec:training}
The goal of training is to learn the warm-start model $h_{\phi}$ and the normalised-space generative model $p'_{\theta}$ required for sampling. This happens in two distinct phases, where we first train $h_{\phi}$ and then $p'_{\theta}$. This modular approach has the following benefits:
\begin{itemize}
    \item $h_\phi$ may be useful as a deterministic model even without $p'_\theta$. For instance, in weather forecasting, both deterministic models and generative models are useful in different contexts \citep{couairon2024archesweather}.
    \item Any existing Gaussian regression model can be used as $h_{\phi}$ without a need for retraining.
    \item Once $h_{\phi}$ is trained, its per-sample outputs can be cached, saving memory and compute when training $p'_{\theta}$.
\end{itemize}

\paragraph{Training the Warm-Start Model}
The goal of the warm-start model is to predict the first two moments of the conditional data distribution $p(X_0 \mid C)$. We do this by training a probabilistic regression model $h_{\phi}$ with parameters $\phi$ using Gaussian negative log-likelihood loss, inspired by conditional neural processes \citep{garnelo_conditional_2018, garnelo_neural_2018}:
\begin{equation}
    \mathcal{L}_{\phi} = - \log p_\phi(X \mid C) = - \log \mathcal{N}(X \mid \hat{\vmu}_C^{(\phi)}, \hat{\vsigma}_C^{(\phi)}).
\end{equation}
Once $h_\phi$ is trained, we freeze its weights.
\paragraph{Training the Generative Model}
Training the normalised-space generative model $p'_\theta$ is best viewed as \textit{transforming the dataset} $\mathcal{D}_{\text{train}}$ into an instance-normalised dataset $\mathcal{D}'_{\text{train}}$ (using $h_\phi$, as outlined in Sec. \ref{sec: conditional normalisation}) and training any off-the-shelf generative model on that modified dataset, which explains why WSD is model-agnostic. The full transformation is shown, for a single training sample, in Alg. \ref{alg:warm-start-training}.

\begin{figure}
\sethlcolor{yellow}
\begin{minipage}[t]{0.49\textwidth}
\begin{savenotes}
\renewcommand*\footnoterule{}
\begin{algorithm}[H]
\caption{Training Step for $p'_{\theta}$}
\begin{algorithmic}[1]
\onehalfspacing
\State \textbf{Input:} $h_{\phi}$, $p'_{\theta}$, $\mathcal{D}_{\text{train}}$, optimizer
\State ($C$, $X_{0}^{(\text{true})}) \sim \mathcal{D}_{\text{train}}$
\State $(\hat{\vmu}_C, \hat{\vsigma}_C) \gets h_{\phi}(C)$
\State $w \sim U[0, 1]$
\State $\vsigma^{\text{norm}}_C \gets w \cdot \max(\hat{\vsigma}_C, 1-w) + (1 - w)\vone$
\State ${X'}_0^{(\text{true})} \gets (X_0^{(\text{true})} - \hat{\vmu}_C) / \hat{\vsigma}_C$
\item[\hl{\textbf{*}:}] \hl{$\mathcal{L} \gets \text{loss}(p'_{\theta}, C, \hat{\vmu}_C, \vsigma^{\text{norm}}_C, w, {X'}_{0}^{(\text{true})})$}
\State $\theta \gets \theta + \text{optimizer}(\nabla_\theta \mathcal{L})$
\end{algorithmic}
\label{alg:warm-start-training}
\end{algorithm}
\end{savenotes}
\end{minipage}
\begin{minipage}[t]{0.49\textwidth}
\begin{algorithm}[H]
\caption{Warm-start Sampling}
\begin{algorithmic}[1]
\onehalfspacing
\State \textbf{Input:} $C, h_{\phi}, p'_{\theta}, [w=1.0]$
\State $(\hat{\vmu}_C, \hat{\vsigma}_C) \gets h_{\phi}(C)$
\State $\vsigma^{\text{norm}}_C \gets w \cdot \max(\hat{\vsigma}_C, 1-w) + (1 - w)\vone$
\State $X'_T \sim \mathcal{N}(0, 1)$
\item[\hl{\textbf{*}:}] \hl{$X'_0 \sim p'_{\theta}(X'_0 \mid X'_T, C, \hat{\vmu}_C, \vsigma^{\text{norm}}_C, w)$}
\State $X_0 \gets X'_0 \cdot \vsigma^{\text{norm}}_C + \hat{\vmu}_C$
\State \textbf{return} $X_0$
\end{algorithmic}
\label{alg:warm-start-sampling}
\end{algorithm}
\end{minipage}
\begin{minipage}[t]{\textwidth}
\footnotesize{\hspace*{0.5cm}\textbf{*}Note that we do not prescribe how to sample from $p'_\theta$, or how its loss is calculated, as WSD is agnostic to the implementation of the generative model.}
\end{minipage}
\end{figure}

\section{Experimental Setup}
\label{sec:experimental-setup}
%While our method can be combined with any advanced generative model, we demonstrate its effectiveness using naive DDPM \citep{ho_denoising_2020} as the generative algorithm. We use the original U-Net architecture \citep{ronneberger_u-net_2015} for both the warm-start model and the generative model, leaving experimentation around how much capacity should be allocated to each model for future work.

%Instead of the original 1000 sampling steps, we use only 10, adjusting the beta schedule to a linear schedule between $\beta_1 = 0.05$ and $\beta_T = 0.5$. These values are significantly larger than in naive DDPM, which is necessary to compensate for the reduced number of sampling steps. Results can likely be improved by tweaking this schedule, which we leave for future work.

%As a baseline, we compare our method to a standard DDPM model (with 1000 timesteps and a linear beta schedule). We also establish a more direct baseline by training a 10-step DDPM-only model, but without warm start. To ensure a fair comparison, this 10-step baseline was also trained using the same aggressive beta schedule as our warm-start-enhanced model.

Across our experiments, we use the Meta Research implementation of flow matching \citep{lipman2024flowmatchingguidecode, lipman_flow_2022} as our baseline, but warm-start models can be combined with any diffusion-based algorithm. We combine this model with the state-of-the-art V3 DPM-Solver \citep{zheng2023dpm_v3}. To make DPM Solver compatible with the flow-matching formalism, we use the equivalence to noise-based diffusion outlined in \citet{gao2025diffusionmeetsflow}. To the best of our knowledge, this is the first time flow matching and DPM Solver are combined, creating a very strong sample-efficient baseline. As flow matching and diffusion can be shown to be different formulations of the same principle \citep{gao2025diffusionmeetsflow, patel2024exploring}, we use both terms interchangeably.

To keep comparisons fair, we use the same architecture for both the baseline and our (warm start) generative models. Additionally, our warm-start model is kept significantly smaller than the generative model, so that one forward pass takes around 1/10th of the time of the generative model. For brevity, we do not include this faster forward pass in our NFE numbers (i.e.~we write NFE=10 instead of NFE=1 fast + 10 slow). For more experiment details, including the model architecture choices, see Appendix \ref{app:experiment-details}.

\section{Image Inpainting}
\begin{figure}
\centering
\includegraphics[width=\textwidth]{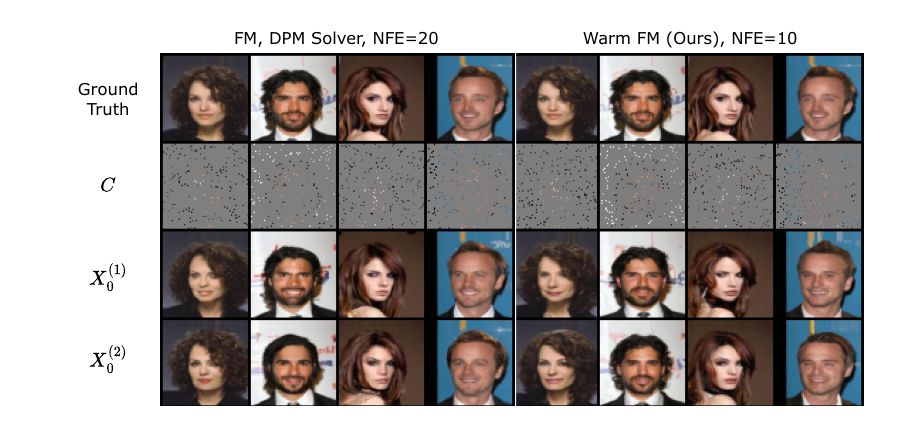}
\caption{Samples $X_0^{(i)}$ generated by standard Flow Matching (NFE=20) and our method (NFE=10).}
\label{fig:celeba-samples}
\end{figure}

In this task, we select a random image from the relevant dataset, and randomly mask out 95\% of the pixels in the image (90\% for CIFAR10 due to the lower resolution). This masked image (as well as the mask itself) is then used as the context data $C$, as shown in Fig. \ref{fig:celeba-samples}.

The models' task is to generate a sample $X_0$ that matches the masked image, i.e.~fills in the missing pixels, while remaining consistent with the unmasked pixels. The entire sampling process is shown in Figure~\ref{fig:inpainting-samples}.

\subsection{Results}

We evaluate our method on the 64x64 CelebA \citep{liu2015faceattributes}, and the 32x32 CIFAR10 \citep{Krizhevsky09learningmultiple} datasets. In both settings, we discard any labels and supplementary information, and only use the masked images (as well as the mask itself) as context data $C$.

As shown in Fig. \ref{fig:celeba-samples}, our method generates realistic samples that are consistent with the unmasked pixels, despite only using NFE=10. These samples are competitive with traditional flow-matching using the DPM solver and NFE=20. Additional samples (including for CIFAR10) can be found in Appendix \ref{app:additional-samples}.

\begin{figure}
\centering
\includegraphics[width=\textwidth]{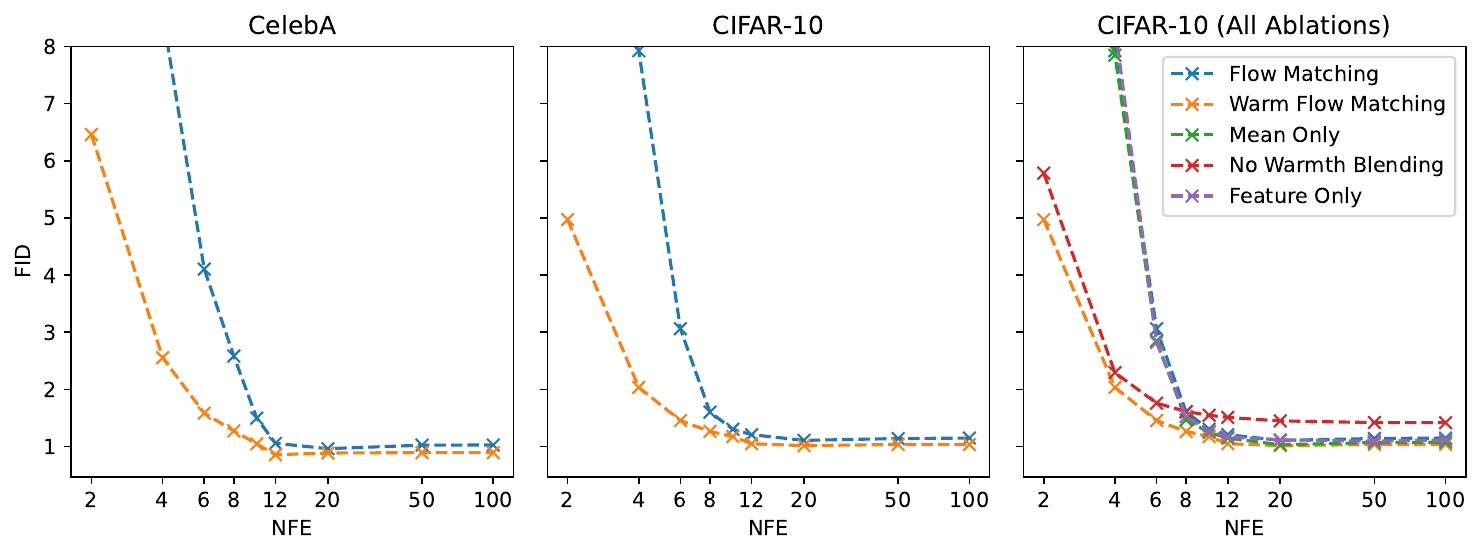}
\caption{Warm-start flow matching substantially outperforms its standard counterpart in the low NFE regime, allowing high-quality samples to be generated in 4-6 function evaluations, and saturating performance in 12.}
\label{fig:inpainting_nfe_vs_fid}
\end{figure}

For quantitative evaluation of perceptual quality, we use the FID (Fréchet inception distance) \citep{heusel2017gans_FID}, computed over 50,000 samples, each evaluated for NFEs between 2 and 100 (Fig. \ref{fig:inpainting_nfe_vs_fid}). Clearly, in the low NFE regime, our method substantially outperforms standard flow matching, able to generate perceptually realistic images using NFE$=4-6$, and saturating performance in 12. Individual samples at different NFE are shown in Appendix \ref{app:additional-samples} (Figs. \ref{fig:cifar10-nfe-progression}, \ref{fig:celeba-nfe-progression}). We also find that our method slightly outperforms the baseline even in the saturated high FID regime. We believe this to be mainly due to the mean subtraction making the modelling task easier, as explained in the mean-only ablation (Sec. \ref{sec:ablations}).

We extensively experiment with various general-purpose and diffusion-specific ODE solvers and integration time discretisations and plot only the best-performing combination at each NFE value. This is generally the midpoint solver using uniform time discretisation for low NFE values (NFE $\leq 5-10$), and the 3rd order DPM Solver using the log-signal-to-noise-ratio time discretisation for NFE $> 5-10$. See Appendix \ref{sec:solvers} for more details.

\subsection{Ablations}
\label{sec:ablations}
All ablations are performed against the CIFAR10 dataset. We do not extend these ablations to other datasets due to computational constraints.
\paragraph{No warmth blending}
Here, we retrain a model without the warmth blending and multi-task training described in Sec. \ref{sec: warmth blending}. This is shown in Fig. \ref{fig:inpainting_nfe_vs_fid} (right, red). Clearly, while the model is still far more NFE-efficient than standard flow matching, it underperforms the blended-warmth model (orange) at all NFE.
\paragraph{Mean-only}
Here, we only use the predicted mean $\hat{\vmu}_C$ for normalisation (equivalent to setting $\hat{\vsigma}_C = \vone$). This is equivalent to training a deterministic (R)MSE model (outputting $\hat{\vmu}_C$) as the shortcut model, and performing diffusion against the residuals. This has shown success in weather forecasting models \citep{couairon2024archesweather, mardani2025residual}. Performance is visualised in Fig. \ref{fig:inpainting_nfe_vs_fid} (right, green). Compared to normal flow-matching, performing diffusion in the residual space improves performance slightly, indicating that this is where our method's high-NFE gains come from, but it performs similarly poorly in the low-NFE regime as standard flow-matching. This also shows that the efficiency gains demonstrated using WSD \emph{heavily depend on the predicted standard deviation}.
\paragraph{Features only}
It could be the case that the increased efficiency comes not from moving $X_T$ closer to $X_0$, but instead from the fact that the generative model $p_\theta$ has access to $\hat{\vmu}_C, \hat{\vsigma}_C$ as inputs. In this case, our method works effectively as a form of feature engineering. We test this by not applying the normalisation, but still providing $\hat{\vmu}_C, \hat{\vsigma}_C$ as inputs to the generative model. As shown in Fig. \ref{fig:inpainting_nfe_vs_fid} (right, purple), this yields no significant improvement over the standard flow-matching baseline, demonstrating that the observed benefits come from the warm-start approach itself, not the additional inputs.

%\begin{table}[t]
%\caption{Evaluation of our method when compared to naive DDPM.}
%\label{table:inpainting-fid}
%\centering
%\begin{tabular}{lccc}
%\toprule
%& & \textbf{CIFAR10 (32x32)} & \textbf{CelebA (64x64)} \\
%\textbf{Method} & \textbf{NFE} & \textbf{FID} $\downarrow$ & \textbf{FID} $\downarrow$ \\
%\midrule
%DDPM\footnote{Note that we were unable to reproduce the FID values reported for CIFAR10 in the original DDPM paper (for unconditional image generation), likely due to a discrepancy in the training setup.} & 1000 & 6.22 & \textbf{2.18} \\
%DDPM & 10 & 15.77 & 5.46 \\
%Warm Start + DDPM (Ours) & 1 + 10 & \textbf{5.27} & 2.19 \\
%\bottomrule
%\end{tabular}
%\end{table}

%\section{ERA5 Wind Inpainting}

\section{ERA5 Wind Forecasting}

In ML-based weather forecasting, the goal is to predict the future weather given the current weather. These systems typically operate on a fixed time interval (e.g.~6 hours). To produce predictions on longer time horizons, the model is applied autoregressively. As the model is trained on \emph{real} weather samples, but deployed autoregressively (using \emph{its own} predictions as inputs), model outputs must be \emph{realistic} weather samples. Otherwise, the model falls increasingly out of distribution when rolled out in time.

\begin{figure}
\centering
\includegraphics[width=\textwidth]{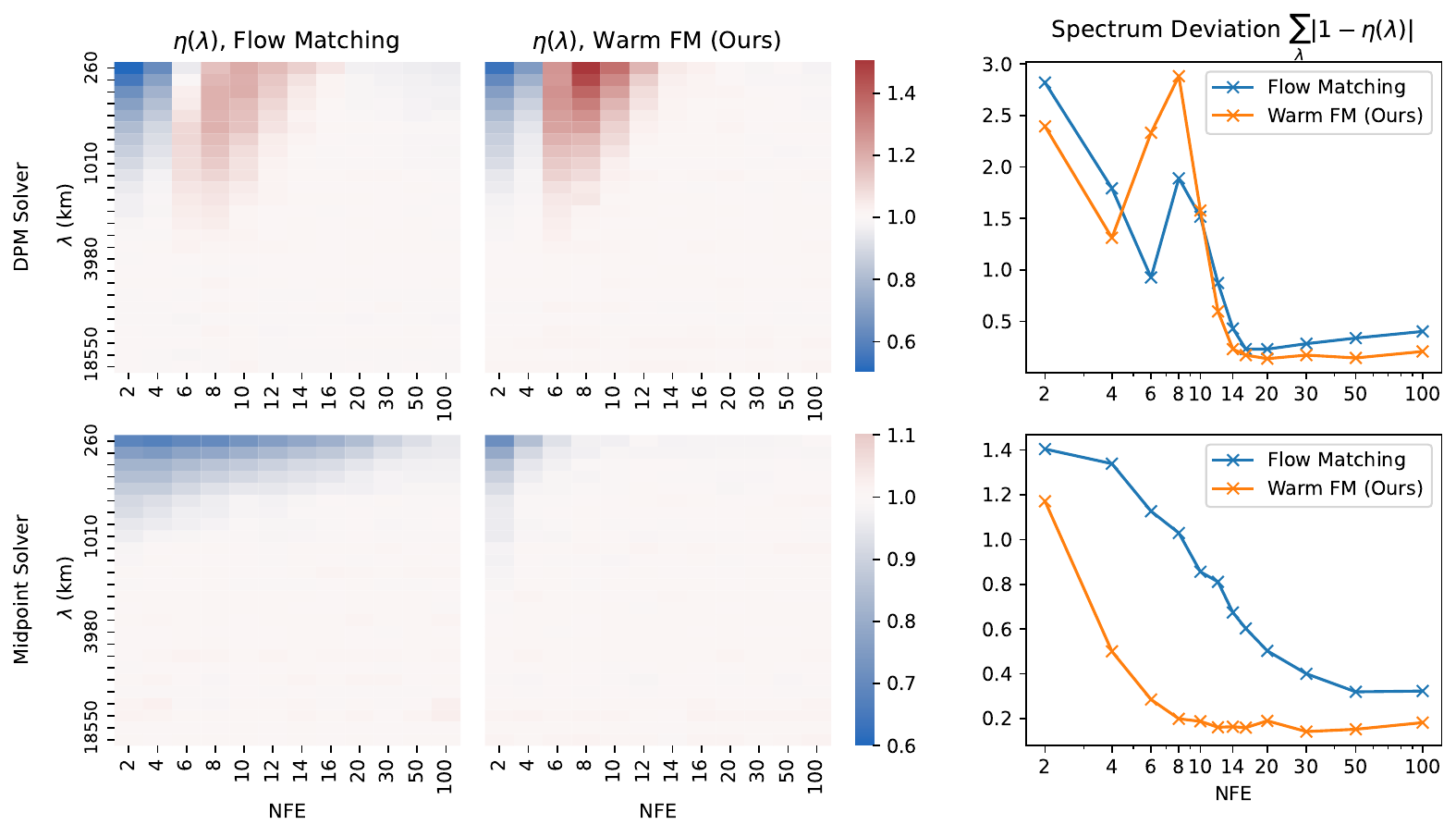}
\caption{\textbf{Left}: The power spectrum ratio, $\eta (\lambda)$, compares the presence of certain wavelengths in the model's predictions to the ground truth: $\eta(\lambda)<1$ (blue) $\implies \lambda$ is under represented, $\eta(\lambda)>1$ (red) $\implies \lambda$ is overrepresented. For low NFE, predictions are blurry. For higher NFE, the generated samples' power spectra align with the ground truth.
\textbf{Right}: By summing the absolute deviations from the ground truth power spectrum $\sum_{\lambda}|1 - \eta(\lambda)|$, we can summarise the power spectrum deviation into a single number at each NFE. \textbf{Top row}: Using DPM Solver, both standard and warm-start flow matching reach their terminal state after $14-20$ NFE. \textbf{Bottom row}: Using the midpoint solver, warm-start flow matching (orange) becomes significantly more efficient than conventional flow matching, needing only $\sim$ NFE=10 to saturate its performance.}
\label{fig:wind_forecasting_eval}
\end{figure}

Existing diffusion-based generative models such as GenCast \citep{price_gencast_2024} have shown good results, but are expensive to run. For instance, a single 15-day forecast with 50 ensemble members at NFE=39 per sample (as performed by \citet{price_gencast_2024}) requires 58,500 forward passes (see Appendix \ref{appendix:calculation-ensemble}), needing $\sim 7$ hours on a single Cloud TPUv5 device \citep{price_gencast_2024}. As shown in Fig. \ref{fig:wind_forecasting_eval}, our method requires only NFE $\approx 10$ per AR Step, reducing compute requirements by $\sim 75\%$.

We emphasise that our goal is not to achieve state-of-the-art forecasting results, but rather to demonstrate that our method can generate realistic weather samples in a fraction of the sampling steps used by current approaches. To do this, we use a lightweight convolutional U-Net \citep{ronneberger_u-net_2015} architecture, and restrict ourselves to only modelling the $u$ and $v$ components of wind 10m above the ground. We also limit ourselves to a spatial resolution of 1.5$^\circ$ (i.e.~240x121 grid points), as provided by the re-gridded ERA5 reanalysis dataset \citep{hersbach_era5_2020}. Our model uses an internal temporal resolution of 6 hours, and is given a snapshot of the current wind fields, and the wind fields 6 hours prior as context data $C$.

\begin{figure}
\centering
\includegraphics[width=\textwidth]{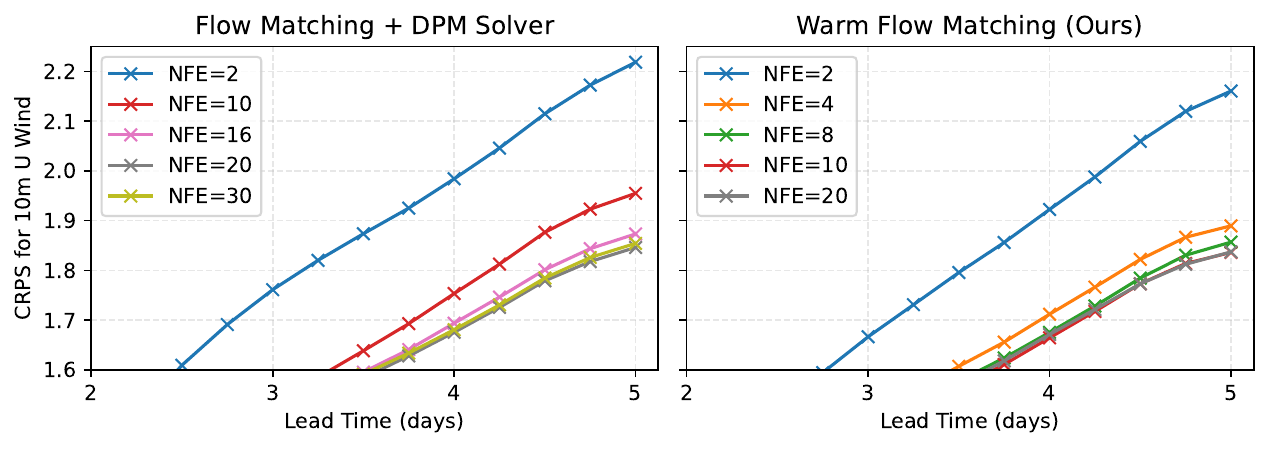}
\caption{Conditional Ranked Probability Score (CRPS) computed over an ensemble of 50 forecast trajectories. With conventional flow matching and DPM Solver (left), the CRPS performance saturates for NFE above $\sim 20$. Using warm-start flow matching (right), performance saturates after NFE=10. The \textit{saturated} performance of both methods is very similar.}
\label{fig:crps}
\end{figure}

\subsection{Results}
\label{sec: weather results}
In the absence of a perceptual accuracy metric like the FID for generated images, we evaluate our models using two commonly used metrics:
\begin{enumerate}
    \item Fig. \ref{fig:crps} shows the Continuous Ranked Probability Score (CRPS) over a 5-day autoregressive forecast using 50 ensemble members. The CRPS is a proper scoring rule which can be considered as a probabilistic generalisation of the mean absolute error.
    \item Fig.~ \ref{fig:wind_forecasting_eval} shows the power spectrum ratio $\eta(\lambda)$. It compares the power of different wavelengths $\lambda$ present in generated samples to the ground truth power. Good samples have $\eta(\lambda) \approx 1 \, \forall \, \lambda$.
\end{enumerate}
In both metrics, standard flow matching (with DPM Solver) shows improvements up to NFE $\approx 20$, whereas WSD saturates performance for NFE above $\approx 10$. Appendix \ref{app:additional-samples} (Fig. \ref{fig:forecast-samples}), visualises forecast trajectories sampled using WSD as well as the ground truth, showing that the warm-start model is capable of generating plausible, yet diverse forecasts.

\section{Conclusion}
In this work, we introduced warm-start models, a widely applicable, easily implemented, and effective method for reducing the number of sampling steps required in conditional generative modelling. By using a simple, deterministic network to predict the initial moments of the conditional data distribution, we effectively reduce the distance the generative process must traverse. This approach is not only orthogonal to and synergistic with existing efficient samplers, but is also simple to implement, allowing it to be freely combined with any generative model. On benchmark tasks like image inpainting and weather forecasting, our approach can generate realistic samples in 4-6 function evaluations, and saturates performance in 10-12, demonstrating a substantial leap in sampling efficiency.

\paragraph{Limitations} The primary limitation of this method lies in the warm-start model's assumption of an uncorrelated Gaussian posterior. This makes it highly effective for tasks with strong conditioning information that lead to a largely unimodal conditional distribution, such as inpainting or weather forecasting. Conversely, its utility is diminished in highly multimodal settings like text-to-image synthesis, where a single Gaussian is an insufficient prior.
Further work is needed to investigate how WSD performs on more multi-modal tasks with weaker conditioning information (e.g.~inpainting with fewer pixels or weather forecasting over longer time intervals). A second limitation is that a separate warm-start model needs to be trained for each experiment and dataset. It may be possible\footnote{In fact, we mistakenly initially used a CIFAR10-trained warm-start model for WSD on CelebA. We found only a small performance loss even though the two datasets are substantially different.} to train a single general-purpose warm-start model (trained e.g.~on Imagenet \citet{deng2009imagenet}) that can be used for any image-related tasks.

\paragraph{Future work} WSD can be made even more efficient and flexible. Predicting a conditional low-rank correlation matrix, instead of only marginal standard deviations, could accelerate the method. Additional speed-ups may come from adapting efficient sampling tricks, like EDM's custom time discretisation \citep{karras_elucidating_2022} or ODE solvers such as DPM-Solver \citet{lu_dpm-solver_2022, lu2025dpmpp, zheng2023dpm_v3}, from standard diffusion to WSD. Finally, WSD opens up the possibility of inference-time compute scaling: by using the uncertainty estimate from the warm-start model to allocate the number of sampling steps (using more for highly uncertain predictions and fewer for confident ones), compute can be dynamically allocated based on need.

These advancements, building upon an already simple, effective, and widely applicable framework, have the potential to make WSD an even more efficient and flexible tool for conditional generation.

\section*{Acknowledgments}
We thank Xianda Sun, Cristiana Diaconu, and Aliaksandra Shysheya for their helpful discussions and feedback on this work. Jonas Scholz is supported by the Cambridge Zero | Marshall Foundation Scholarship. Richard E. Turner is supported by Google, Amazon, ARM, Improbable and an EPSRC Prosperity Partnership (EP/T005386/1) and the EPSRC Probabilistic AI Hub (EP/Y028783/1).

\section*{Reproducibility Statement}
We make our method reproducible by outlining the method in Sec. \ref{sec:warm-start-models}, providing the broad experimental setup in Sec. \ref{sec:experimental-setup}, providing more details in Appendix \ref{app:experiment-details}, and also providing the anonymised source code for review. After anonymous peer review, we will make the source code available on GitHub.

\bibliography{main}

\begin{thebibliography}{33}
\providecommand{\natexlab}[1]{#1}
\providecommand{\url}[1]{\texttt{#1}}
\expandafter\ifx\csname urlstyle\endcsname\relax
  \providecommand{\doi}[1]{doi: #1}\else
  \providecommand{\doi}{doi: \begingroup \urlstyle{rm}\Url}\fi

\bibitem[Alet et~al.(2025)Alet, Price, El-Kadi, Masters, Markou, Andersson, Stott, Lam, Willson, Sanchez-Gonzalez, et~al.]{alet2025skillful}
Ferran Alet, Ilan Price, Andrew El-Kadi, Dominic Masters, Stratis Markou, Tom~R Andersson, Jacklynn Stott, Remi Lam, Matthew Willson, Alvaro Sanchez-Gonzalez, et~al.
\newblock Skillful joint probabilistic weather forecasting from marginals.
\newblock \emph{arXiv preprint arXiv:2506.10772}, 2025.

\bibitem[Chen(2018)]{torchdiffeq}
Ricky T.~Q. Chen.
\newblock torchdiffeq, 2018.
\newblock URL \url{https://github.com/rtqichen/torchdiffeq}.

\bibitem[Couairon et~al.(2024)Couairon, Singh, Charantonis, Lessig, and Monteleoni]{couairon2024archesweather}
Guillaume Couairon, Renu Singh, Anastase Charantonis, Christian Lessig, and Claire Monteleoni.
\newblock Archesweather \& archesweathergen: a deterministic and generative model for efficient ml weather forecasting.
\newblock \emph{arXiv preprint arXiv:2412.12971}, 2024.

\bibitem[Deng et~al.(2009)Deng, Dong, Socher, Li, Li, and Fei-Fei]{deng2009imagenet}
Jia Deng, Wei Dong, Richard Socher, Li-Jia Li, Kai Li, and Li~Fei-Fei.
\newblock Imagenet: A large-scale hierarchical image database.
\newblock In \emph{2009 IEEE conference on computer vision and pattern recognition}, pp.\  248--255. Ieee, 2009.

\bibitem[Gao et~al.(2024)Gao, Hoogeboom, Heek, Bortoli, Murphy, and Salimans]{gao2025diffusionmeetsflow}
Ruiqi Gao, Emiel Hoogeboom, Jonathan Heek, Valentin~De Bortoli, Kevin~P. Murphy, and Tim Salimans.
\newblock Diffusion meets flow matching: Two sides of the same coin.
\newblock 2024.
\newblock URL \url{https://diffusionflow.github.io/}.

\bibitem[Garnelo et~al.(2018{\natexlab{a}})Garnelo, Rosenbaum, Maddison, Ramalho, Saxton, Shanahan, Teh, Rezende, and Eslami]{garnelo_conditional_2018}
Marta Garnelo, Dan Rosenbaum, Chris~J. Maddison, Tiago Ramalho, David Saxton, Murray Shanahan, Yee~Whye Teh, Danilo~J. Rezende, and S.~M.~Ali Eslami.
\newblock Conditional neural processes, 2018{\natexlab{a}}.
\newblock URL \url{http://arxiv.org/abs/1807.01613}.

\bibitem[Garnelo et~al.(2018{\natexlab{b}})Garnelo, Schwarz, Rosenbaum, Viola, Rezende, Eslami, and Teh]{garnelo_neural_2018}
Marta Garnelo, Jonathan Schwarz, Dan Rosenbaum, Fabio Viola, Danilo~J. Rezende, S.~M.~Ali Eslami, and Yee~Whye Teh.
\newblock Neural processes, 2018{\natexlab{b}}.
\newblock URL \url{http://arxiv.org/abs/1807.01622}.

\bibitem[Goodfellow et~al.(2020)Goodfellow, Pouget-Abadie, Mirza, Xu, Warde-Farley, Ozair, Courville, and Bengio]{goodfellow2020generative}
Ian Goodfellow, Jean Pouget-Abadie, Mehdi Mirza, Bing Xu, David Warde-Farley, Sherjil Ozair, Aaron Courville, and Yoshua Bengio.
\newblock Generative adversarial networks.
\newblock \emph{Communications of the ACM}, 63\penalty0 (11):\penalty0 139--144, 2020.

\bibitem[Hersbach et~al.(2020)Hersbach, Bell, Berrisford, Hirahara, Horányi, Muñoz-Sabater, Nicolas, Peubey, Radu, Schepers, Simmons, Soci, Abdalla, Abellan, Balsamo, Bechtold, Biavati, Bidlot, Bonavita, De~Chiara, Dahlgren, Dee, Diamantakis, Dragani, Flemming, Forbes, Fuentes, Geer, Haimberger, Healy, Hogan, Hólm, Janisková, Keeley, Laloyaux, Lopez, Lupu, Radnoti, de~Rosnay, Rozum, Vamborg, Villaume, and Thépaut]{hersbach_era5_2020}
Hans Hersbach, Bill Bell, Paul Berrisford, Shoji Hirahara, András Horányi, Joaquín Muñoz-Sabater, Julien Nicolas, Carole Peubey, Raluca Radu, Dinand Schepers, Adrian Simmons, Cornel Soci, Saleh Abdalla, Xavier Abellan, Gianpaolo Balsamo, Peter Bechtold, Gionata Biavati, Jean Bidlot, Massimo Bonavita, Giovanna De~Chiara, Per Dahlgren, Dick Dee, Michail Diamantakis, Rossana Dragani, Johannes Flemming, Richard Forbes, Manuel Fuentes, Alan Geer, Leo Haimberger, Sean Healy, Robin~J. Hogan, Elías Hólm, Marta Janisková, Sarah Keeley, Patrick Laloyaux, Philippe Lopez, Cristina Lupu, Gabor Radnoti, Patricia de~Rosnay, Iryna Rozum, Freja Vamborg, Sebastien Villaume, and Jean-Noël Thépaut.
\newblock The {ERA}5 global reanalysis.
\newblock 146\penalty0 (730):\penalty0 1999--2049, 2020.
\newblock ISSN 1477-870X.
\newblock \doi{10.1002/qj.3803}.
\newblock URL \url{https://onlinelibrary.wiley.com/doi/abs/10.1002/qj.3803}.
\newblock \_eprint: https://rmets.onlinelibrary.wiley.com/doi/pdf/10.1002/qj.3803.

\bibitem[Heusel et~al.(2017)Heusel, Ramsauer, Unterthiner, Nessler, and Hochreiter]{heusel2017gans_FID}
Martin Heusel, Hubert Ramsauer, Thomas Unterthiner, Bernhard Nessler, and Sepp Hochreiter.
\newblock Gans trained by a two time-scale update rule converge to a local nash equilibrium.
\newblock \emph{Advances in neural information processing systems}, 30, 2017.

\bibitem[Ho et~al.(2020)Ho, Jain, and Abbeel]{ho_denoising_2020}
Jonathan Ho, Ajay Jain, and Pieter Abbeel.
\newblock Denoising diffusion probabilistic models, 2020.
\newblock URL \url{http://arxiv.org/abs/2006.11239}.

\bibitem[Ho et~al.(2022)Ho, Chan, Saharia, Whang, Gao, Gritsenko, Kingma, Poole, Norouzi, Fleet, and Salimans]{ho_imagen_2022}
Jonathan Ho, William Chan, Chitwan Saharia, Jay Whang, Ruiqi Gao, Alexey Gritsenko, Diederik~P. Kingma, Ben Poole, Mohammad Norouzi, David~J. Fleet, and Tim Salimans.
\newblock Imagen video: High definition video generation with diffusion models, 2022.
\newblock URL \url{http://arxiv.org/abs/2210.02303}.

\bibitem[Hoogeboom et~al.(2022)Hoogeboom, Satorras, Vignac, and Welling]{3d_molecule_diffusion}
Emiel Hoogeboom, V\'{\i}ctor~Garcia Satorras, Cl{\'e}ment Vignac, and Max Welling.
\newblock Equivariant diffusion for molecule generation in 3{D}.
\newblock In Kamalika Chaudhuri, Stefanie Jegelka, Le~Song, Csaba Szepesvari, Gang Niu, and Sivan Sabato (eds.), \emph{Proceedings of the 39th International Conference on Machine Learning}, volume 162 of \emph{Proceedings of Machine Learning Research}, pp.\  8867--8887. PMLR, 17--23 Jul 2022.
\newblock URL \url{https://proceedings.mlr.press/v162/hoogeboom22a.html}.

\bibitem[Karras et~al.(2022)Karras, Aittala, Aila, and Laine]{karras_elucidating_2022}
Tero Karras, Miika Aittala, Timo Aila, and Samuli Laine.
\newblock Elucidating the design space of diffusion-based generative models, 2022.
\newblock URL \url{http://arxiv.org/abs/2206.00364}.

\bibitem[Kong et~al.(2021)Kong, Ping, Huang, Zhao, and Catanzaro]{kong_diffwave_2021}
Zhifeng Kong, Wei Ping, Jiaji Huang, Kexin Zhao, and Bryan Catanzaro.
\newblock {DiffWave}: A versatile diffusion model for audio synthesis, 2021.
\newblock URL \url{http://arxiv.org/abs/2009.09761}.

\bibitem[Krizhevsky(2009)]{Krizhevsky09learningmultiple}
Alex Krizhevsky.
\newblock Learning multiple layers of features from tiny images.
\newblock Technical report, University of Toronto, 2009.
\newblock URL \url{https://www.cs.toronto.edu/~kriz/cifar.html}.

\bibitem[Lang et~al.(2024)Lang, Alexe, Clare, Roberts, Adewoyin, Bouall{\`e}gue, Chantry, Dramsch, Dueben, Hahner, et~al.]{lang2024aifscrps}
Simon Lang, Mihai Alexe, Mariana~CA Clare, Christopher Roberts, Rilwan Adewoyin, Zied~Ben Bouall{\`e}gue, Matthew Chantry, Jesper Dramsch, Peter~D Dueben, Sara Hahner, et~al.
\newblock Aifs-crps: ensemble forecasting using a model trained with a loss function based on the continuous ranked probability score.
\newblock \emph{arXiv preprint arXiv:2412.15832}, 2024.

\bibitem[Lipman et~al.(2022)Lipman, Chen, Ben-Hamu, Nickel, and Le]{lipman_flow_2022}
Yaron Lipman, Ricky T.~Q. Chen, Heli Ben-Hamu, Maximilian Nickel, and Matthew Le.
\newblock Flow matching for generative modeling.
\newblock 2022.
\newblock URL \url{https://openreview.net/forum?id=PqvMRDCJT9t}.

\bibitem[Lipman et~al.(2024)Lipman, Havasi, Holderrieth, Shaul, Le, Karrer, Chen, Lopez-Paz, Ben-Hamu, and Gat]{lipman2024flowmatchingguidecode}
Yaron Lipman, Marton Havasi, Peter Holderrieth, Neta Shaul, Matt Le, Brian Karrer, Ricky T.~Q. Chen, David Lopez-Paz, Heli Ben-Hamu, and Itai Gat.
\newblock Flow matching guide and code, 2024.
\newblock URL \url{https://arxiv.org/abs/2412.06264}.

\bibitem[Liu et~al.(2015)Liu, Luo, Wang, and Tang]{liu2015faceattributes}
Ziwei Liu, Ping Luo, Xiaogang Wang, and Xiaoou Tang.
\newblock Deep learning face attributes in the wild.
\newblock In \emph{Proceedings of International Conference on Computer Vision (ICCV)}, December 2015.

\bibitem[Loshchilov \& Hutter(2019)Loshchilov and Hutter]{loshchilov_adamw}
Ilya Loshchilov and Frank Hutter.
\newblock Decoupled weight decay regularization, 2019.
\newblock URL \url{http://arxiv.org/abs/1711.05101}.

\bibitem[Lu et~al.(2022)Lu, Zhou, Bao, Chen, Li, and Zhu]{lu_dpm-solver_2022}
Cheng Lu, Yuhao Zhou, Fan Bao, Jianfei Chen, Chongxuan Li, and Jun Zhu.
\newblock {DPM}-solver: A fast {ODE} solver for diffusion probabilistic model sampling in around 10 steps, 2022.
\newblock URL \url{http://arxiv.org/abs/2206.00927}.

\bibitem[Lu et~al.(2025)Lu, Zhou, Bao, Chen, Li, and Zhu]{lu2025dpmpp}
Cheng Lu, Yuhao Zhou, Fan Bao, Jianfei Chen, Chongxuan Li, and Jun Zhu.
\newblock Dpm-solver++: Fast solver for guided sampling of diffusion probabilistic models.
\newblock \emph{Machine Intelligence Research}, pp.\  1--22, 2025.

\bibitem[Mardani et~al.(2025)Mardani, Brenowitz, Cohen, Pathak, Chen, Liu, Vahdat, Nabian, Ge, Subramaniam, et~al.]{mardani2025residual}
Morteza Mardani, Noah Brenowitz, Yair Cohen, Jaideep Pathak, Chieh-Yu Chen, Cheng-Chin Liu, Arash Vahdat, Mohammad~Amin Nabian, Tao Ge, Akshay Subramaniam, et~al.
\newblock Residual corrective diffusion modeling for km-scale atmospheric downscaling.
\newblock \emph{Communications Earth \& Environment}, 6\penalty0 (1):\penalty0 124, 2025.

\bibitem[Patel et~al.(2024)Patel, DeLoye, and Mathias]{patel2024exploring}
Zeeshan Patel, James DeLoye, and Lance Mathias.
\newblock Exploring diffusion and flow matching under generator matching.
\newblock \emph{arXiv preprint arXiv:2412.11024}, 2024.

\bibitem[Price et~al.(2024)Price, Sanchez-Gonzalez, Alet, Andersson, El-Kadi, Masters, Ewalds, Stott, Mohamed, Battaglia, Lam, and Willson]{price_gencast_2024}
Ilan Price, Alvaro Sanchez-Gonzalez, Ferran Alet, Tom~R. Andersson, Andrew El-Kadi, Dominic Masters, Timo Ewalds, Jacklynn Stott, Shakir Mohamed, Peter Battaglia, Remi Lam, and Matthew Willson.
\newblock {GenCast}: Diffusion-based ensemble forecasting for medium-range weather, 2024.
\newblock URL \url{http://arxiv.org/abs/2312.15796}.

\bibitem[Ronneberger et~al.(2015)Ronneberger, Fischer, and Brox]{ronneberger_u-net_2015}
Olaf Ronneberger, Philipp Fischer, and Thomas Brox.
\newblock U-net: Convolutional networks for biomedical image segmentation, 2015.
\newblock URL \url{http://arxiv.org/abs/1505.04597}.

\bibitem[Shu et~al.(2023)Shu, Li, and Farimani]{shu2023physics}
Dule Shu, Zijie Li, and Amir~Barati Farimani.
\newblock A physics-informed diffusion model for high-fidelity flow field reconstruction.
\newblock \emph{Journal of Computational Physics}, 478:\penalty0 111972, 2023.

\bibitem[Song et~al.(2022)Song, Meng, and Ermon]{song_denoising_2022}
Jiaming Song, Chenlin Meng, and Stefano Ermon.
\newblock Denoising diffusion implicit models, 2022.
\newblock URL \url{http://arxiv.org/abs/2010.02502}.

\bibitem[Song et~al.(2020)Song, Sohl-Dickstein, Kingma, Kumar, Ermon, and Poole]{song_score-based_2020}
Yang Song, Jascha Sohl-Dickstein, Diederik~P. Kingma, Abhishek Kumar, Stefano Ermon, and Ben Poole.
\newblock Score-based generative modeling through stochastic differential equations.
\newblock 2020.
\newblock URL \url{https://openreview.net/forum?id=PxTIG12RRHS&utm_campaign=NLP%20News&utm_medium=email&utm_source=Revue%20newsletter}.

\bibitem[Song et~al.(2023)Song, Dhariwal, Chen, and Sutskever]{song2023consistency}
Yang Song, Prafulla Dhariwal, Mark Chen, and Ilya Sutskever.
\newblock Consistency models.
\newblock 2023.

\bibitem[Xu et~al.(2022)Xu, Yu, Song, Shi, Ermon, and Tang]{xugeodiff}
Minkai Xu, Lantao Yu, Yang Song, Chence Shi, Stefano Ermon, and Jian Tang.
\newblock Geodiff: A geometric diffusion model for molecular conformation generation.
\newblock In \emph{International Conference on Learning Representations}, 2022.

\bibitem[Zheng et~al.(2023)Zheng, Lu, Chen, and Zhu]{zheng2023dpm_v3}
Kaiwen Zheng, Cheng Lu, Jianfei Chen, and Jun Zhu.
\newblock Dpm-solver-v3: Improved diffusion ode solver with empirical model statistics.
\newblock In \emph{Thirty-seventh Conference on Neural Information Processing Systems}, 2023.

\end{thebibliography}
\bibliographystyle{iclr2026_conference}

\appendix

\section{LLM Declaration}
We used LLMs to assist with writing code and iterating on the language in the final paper.

\section{Experimental Details}
\label{app:experiment-details}

\paragraph{Datasets}
All datasets are normalised. For images, we normalise values to lie between [-1, 1]. For the weather forecasting task, we apply a per-variable normalisation to ensure zero-mean and unit variance.

\paragraph{Warm-start model} We parameterise $h_\phi$ as a lightweight U-Net \citep{ronneberger_u-net_2015} with [64, 128, 256] channels per block and 2 layers per block. We use attention in the second and third block. For the weather forecasting task, we instead use [128, 256, 512] channels, but no attention (as the resolution is much higher, and attention would become computationally expensive).
We train the warm-start model until convergence ( $\approx 2$ million steps) at a batch size of 32 using AdamW at a constant learning rate of 1e-4 (and using default weight decay and betas). We clip the predicted standard deviation at 0.01 to stabilise training and avoid numerical instability when performing normalisation. For the inpainting tasks, we train the model over a range of inpainting tasks, ranging from 3\% of pixels to 10\% of pixels for CelebA, and 5\% of pixels to 20\% of pixels for CIFAR10.

\paragraph{Generative model} 
We choose to follow \citet{lipman2024flowmatchingguidecode} in the model architecture and training procedure for $p'_\theta$. In particular, we use the same U-Net architecture, and train it using the AdamW optimiser \citep{loshchilov_adamw} with a constant learning rate of 1e-4, and with $\beta_1 = 0.9, \beta_2 = 0.95$. We train using an effective batch size of 512 until convergence ($\approx 1.5$ million steps). We condition the model on the diffusion timestep $t$ and the warmth $w$ by computing embeddings and using them to shift and scale features after normalisation. We use exponential moving average (EMA) weight smoothing with a rate of 0.999. We clip gradients with norms above 3.0.
For the weather forecasting experiment, we use a batch size of 4, also training until convergence.

For full details, we refer to the provided source code, and particularly the configuration files.

\subsection{Best Solvers}
\begin{figure}
\centering
\includegraphics[width=\textwidth]{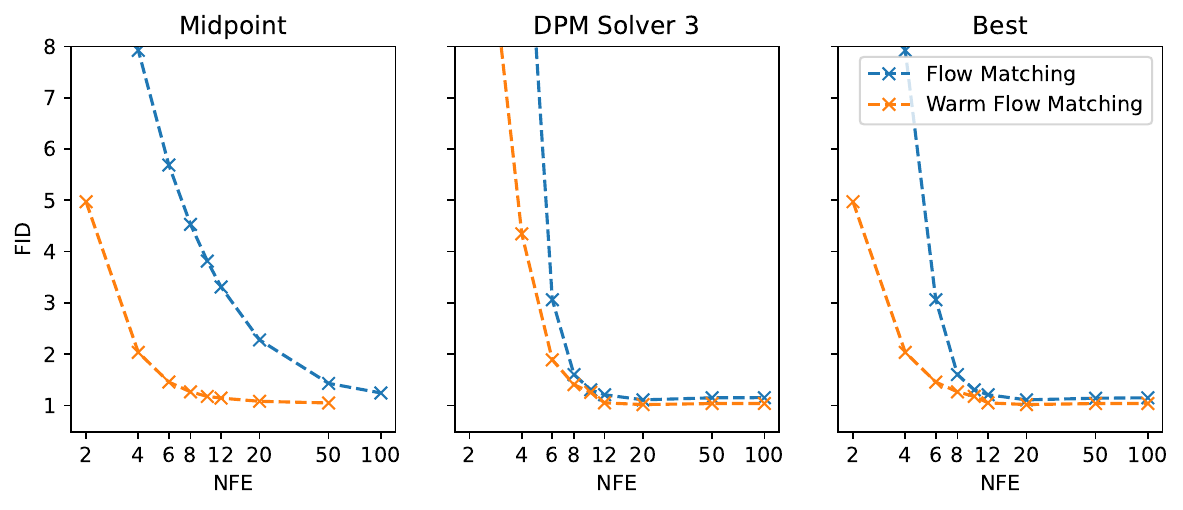}
\caption{On CIFAR10, warm-start diffusion substantially outperforms its standard ``cold" counterpart in the low NFE regime, allowing high-quality samples to be generated in 6 function evaluations, and saturating performance in 12. The performance gap is very pronounced for the simpler midpoint solver (left). Using DPM Solver makes standard flow matching more competitive (middle), but when using the best solver at each NFE, the performance gain }
\label{fig:inpainting_nfe_vs_fid_cifar10}
\end{figure}
\label{sec:solvers}
When comparing results (e.g.~in Fig. \ref{fig:inpainting_nfe_vs_fid}), we evaluate each data point using a combination of ODE solvers and time discretisations. We find that in the very low NFE regime ($\leq 5$ for standard diffusion, $\leq 10$ for warm start diffusion), the best results are achieved using the midpoint ODE solver using a uniform time discretisation. For higher NFE, we find that the 3rd order DPM Solver using a log signal-to-noise ratio time discretisation achieves the best results. For very high NFE ($> 50$), we sometimes find that performance slightly degrades using DPM Solvers.

We tested an extensive selection of ODE solvers and time discretisations. Specifically, we test all fixed step solvers available in the torchdiffeq library \citep{torchdiffeq}, and the following time discretisation schemes:
\begin{itemize}
    \item Uniform in time
    \item Quadratic in time
    \item Log signal-to-noise ratio
    \item The EDM discretisation proposed in \citet{karras_elucidating_2022}.
\end{itemize}
We find that these choices have a large impact on sample efficiency, and we also find that warm-start diffusion is more robust to suboptimal choices than standard diffusion. A selection of results produced by different solvers is shown in Fig. \ref{fig:inpainting_nfe_vs_fid_cifar10}.

\section{NFE Calculation Weather Forecasting}
\label{appendix:calculation-ensemble}
A 15-day forecast with 50 ensemble members at NFE=39 per sample (as performed by \citet{price_gencast_2024}) requires: \vspace{0pt}
\begin{equation}
    50 \text{ Ens. Members} \times \frac{15 \text{ Days}}{\text{Ens. Member}} \times \frac{2 \text{ AR Steps}}{\text{ Day}} \times \frac{39 \text{ Fwd. Passes}}{\text{AR Step}} = 58,500 \text{ Fwd. Passes.}
\end{equation}

\section{Additional Samples}
\label{app:additional-samples}
\begin{figure}
\centering
\includegraphics[width=\textwidth]{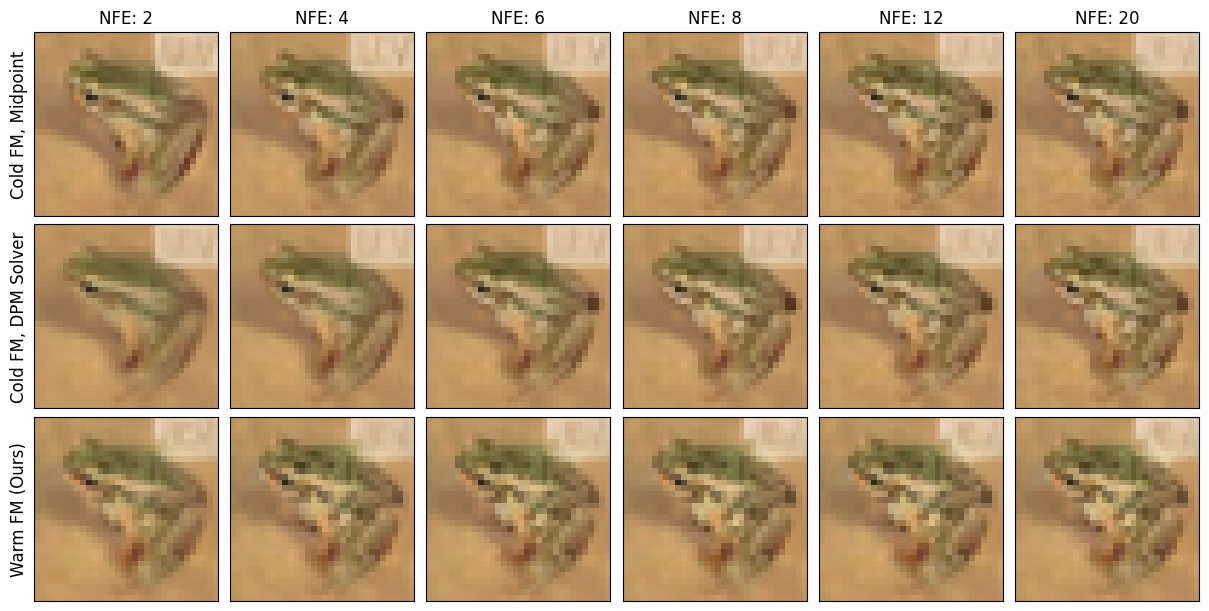}
\caption{Evaluating samples drawn from the same context and same random noise at different NFE. While standard diffusion produces blurry samples for NFE=2-4, warm diffusion is already able to include high-frequency details. For warm diffusion, past NFE $\sim 4-6$, the samples do not visibly change. For standard diffusion, even when using DPM Solver, additional details in the frog's skin texture appear for NFE up to $\sim 12-20$.}
\label{fig:cifar10-nfe-progression}
\end{figure}

\begin{figure}
\centering
\includegraphics[width=\textwidth]{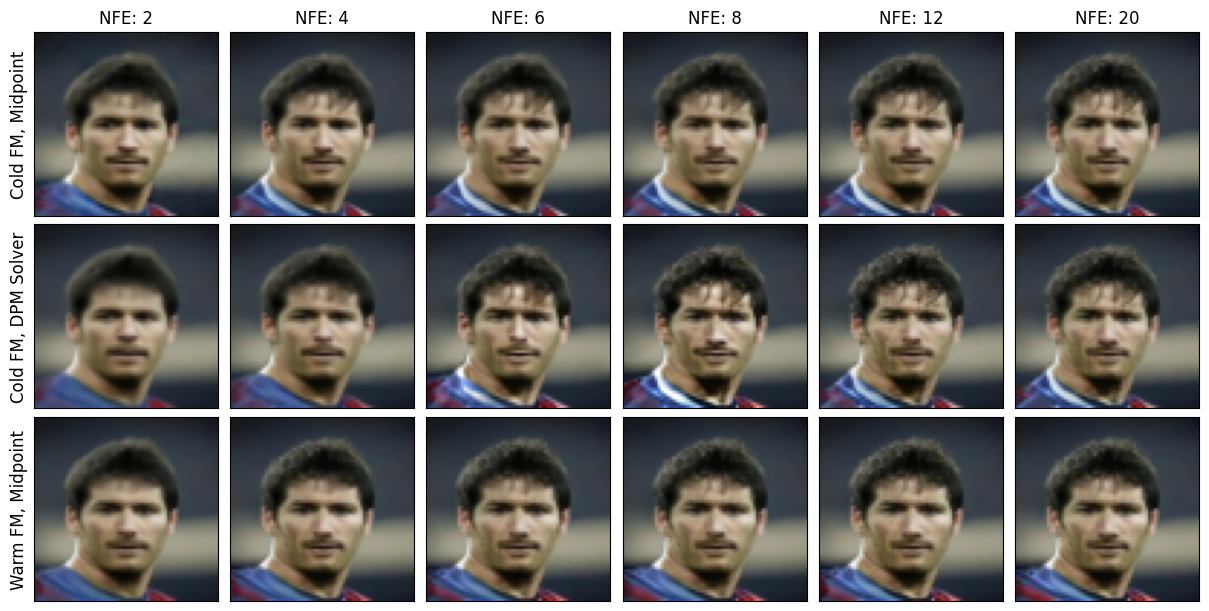}
\caption{Like Fig. \ref{fig:cifar10-nfe-progression} but for the CelebA dataset.}
\label{fig:celeba-nfe-progression}
\end{figure}

We compare warm-start diffusion to standard diffusion qualitatively at different NFE in Figs. \ref{fig:cifar10-nfe-progression} (CIFAR10) and \ref{fig:celeba-nfe-progression} (CelebA), showing that details appear for lower NFE values when using WSD.

\begin{figure}
\centering
\includegraphics[width=\linewidth]{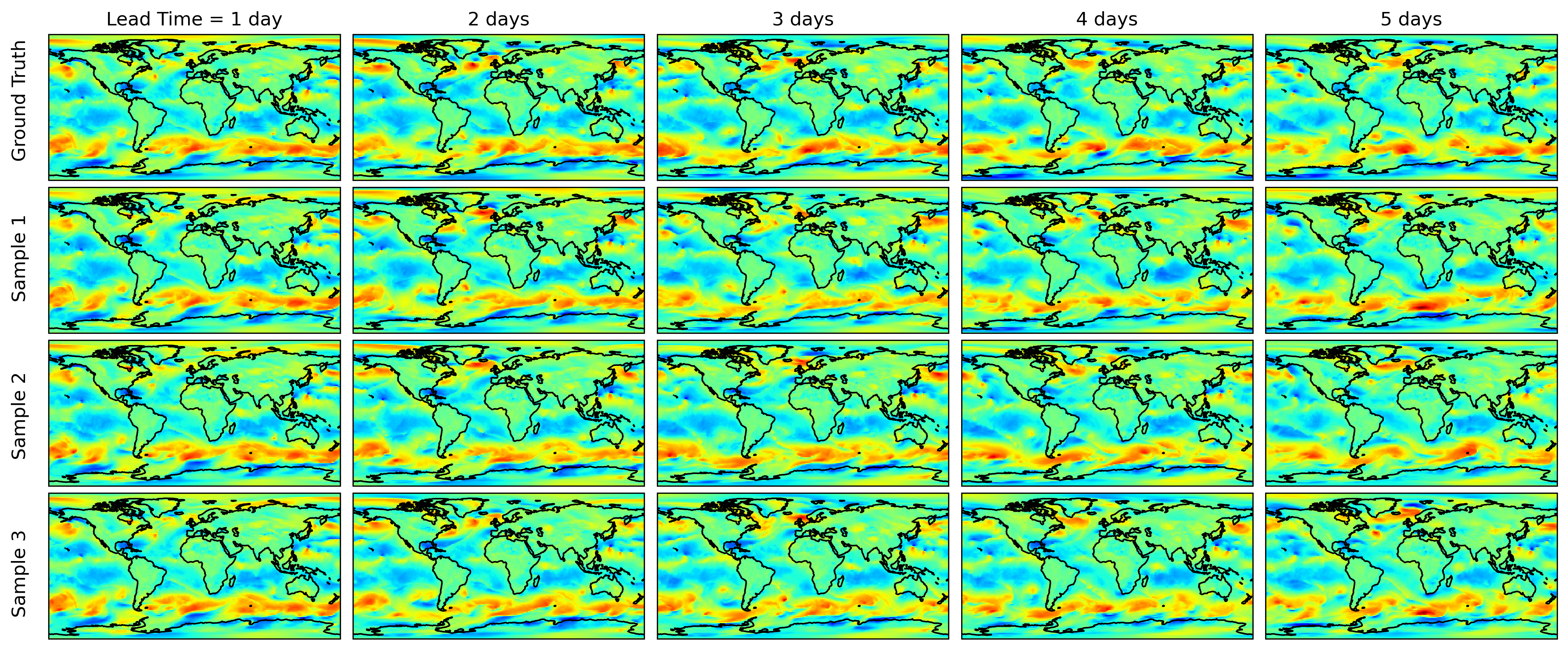}
\caption{Autoregressive forecast trajectories for the U-component of wind at 10m, generated using NFE=10. \textbf{Top row}: Ground truth ERA5 data. \textbf{Bottom three rows}: Four independent forecast samples generated by our method (NFE=11 per 6-hour step), starting from the same initial conditions. The forecasts remain plausible and diverge from each other, demonstrating the model's ability to produce a probabilistic ensemble.}
\label{fig:forecast-samples}
\end{figure}

\begin{figure}
\centering
\includegraphics[width=\textwidth]{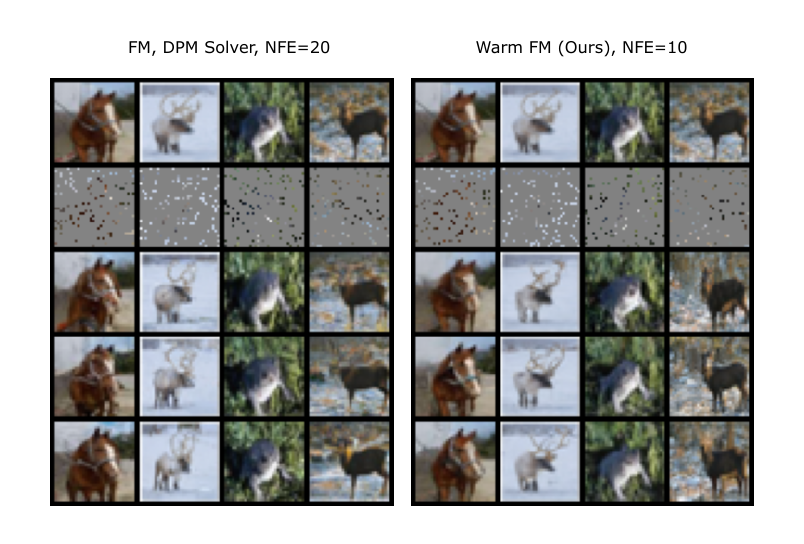}
\caption{Like Fig. \ref{fig:celeba-samples} but for CIFAR10.}
\label{fig:cifar10-samples}
\end{figure}

\begin{figure}
\centering
\includegraphics[width=\textwidth]{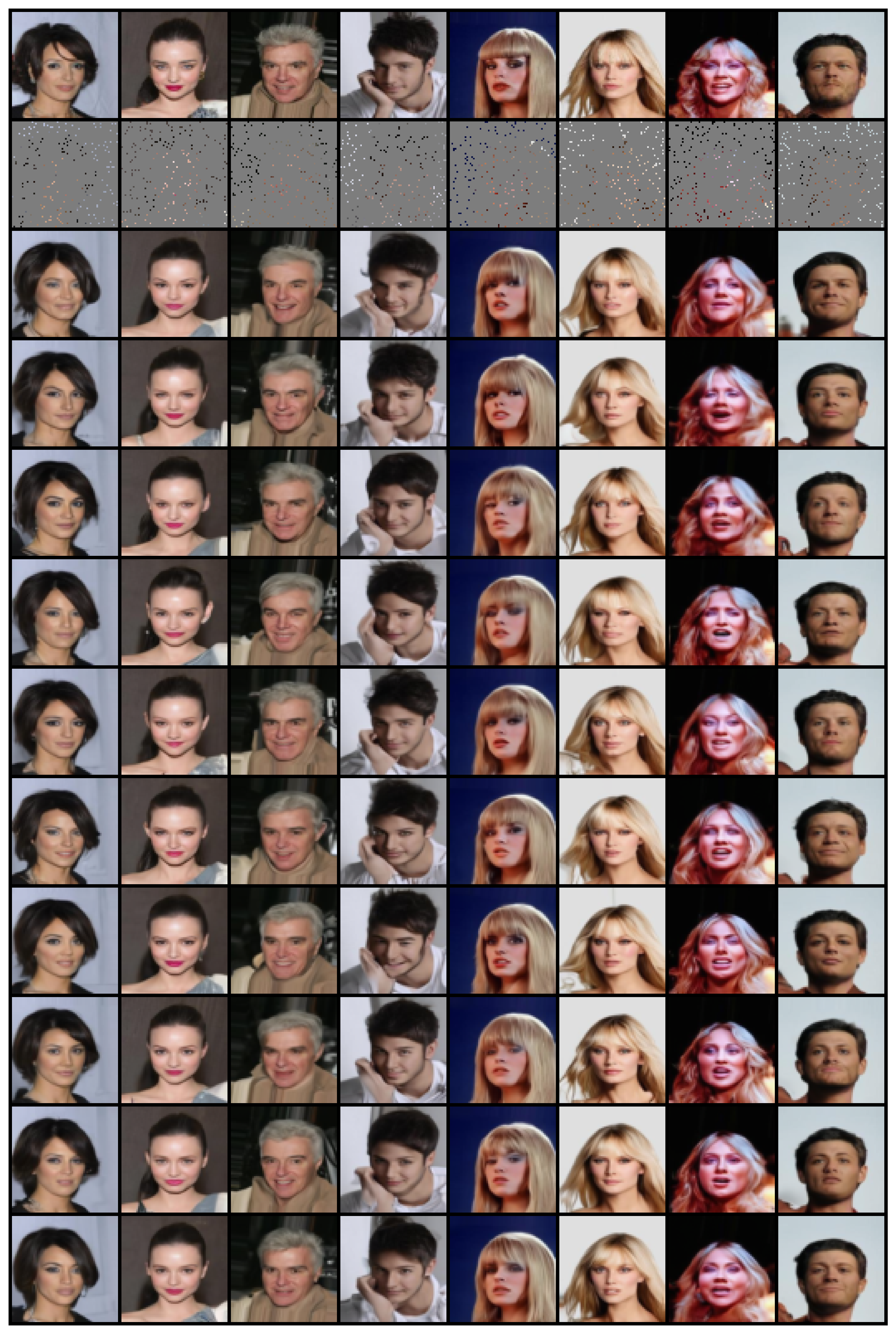}
\caption{Additional CelebA inpainting samples.}
\label{fig:more-celeba}
\end{figure}

In Fig. \ref{fig:forecast-samples}, we show a 3-member ensemble of 5-day wind forecasting trajectories in Fig. \ref{fig:forecast-samples}. In Figs. \ref{fig:cifar10-samples} and \ref{fig:more-celeba}, we provide additional samples for CIFAR10 and CelebA inpainting respectively.

\end{document}